\documentclass[conference]{IEEEtran}

\makeatletter
\def\ps@IEEEtitlepagestyle{%
  \def\@oddfoot{\mycopyrightnotice}%
  \def\@evenfoot{}%
}
\def\mycopyrightnotice{%
  \parbox{\textwidth}{\scriptsize \copyright~2026 IEEE. Personal use of this material is permitted. Permission from IEEE must be obtained for all other uses, in any current or future media, including reprinting/republishing this material for advertising or promotional purposes, creating new collective works, for resale or redistribution to servers or lists, or reuse of any copyrighted component of this work in other works.}%
  \gdef\mycopyrightnotice{}%
}
\makeatother

\ifCLASSINFOpdf
\else
\fi

\usepackage{stfloats}
\usepackage{graphicx}
\usepackage{amsmath}
\usepackage{booktabs}
\usepackage{pifont}
\usepackage{float}

\usepackage[skip=4pt]{caption}
\begin{document}
\title{Weakly Supervised Fish Segmentation through
Point-Supervised Tracking and Motion-Informed
Language}

\author{
\IEEEauthorblockN{Vedanth Kallakuri}
\IEEEauthorblockA{
University of Michigan\\
Ann Arbor, Michigan\\
}
\and
\IEEEauthorblockN{Katherine A. Skinner}
\IEEEauthorblockA{
University of Michigan\\
Ann Arbor, Michigan\\
}}

\maketitle

\begin{abstract}
Understanding the properties of fish populations is increasingly important. Analyzing fish length, width, and mass is integral to maintaining healthy stock in the aquaculture industry, while counting wild fish populations supports marine conservation and coastal reef restoration efforts. Deep learning-based segmentation models offer a promising means of obtaining such information at scale. Most of these models, however, require costly and time-consuming per-pixel segmentation labels for training, which can take up to two minutes per fish to annotate manually. %
To address this inefficiency, we propose a weakly supervised segmentation pipeline that requires only point-labeled data (one click per fish) to produce accurate segmentation masks. Our pipeline consists of four stages. First, a point-supervised tracker predicts point annotations for each frame. These points are then converted into motion-informed language expressions, which guide a language-conditioned segmentation model (LMPM++) %
to capture temporal consistency and produce candidate segmentation masks. Finally, optical flow and background subtraction are applied as post-processing filters to suppress noise from incorrect segmentations. Experiments on the DeepFish dataset %
show improvement in foreground mean intersection over union (mIoU) as compared to baseline models.
\end{abstract}

\IEEEpeerreviewmaketitle

\section{Introduction}

Gaining insight into the properties of marine environments has become increasingly important with the growth of maritime technologies such as autonomous underwater vehicles (AUVs) and advanced computer vision tools \cite{garcia2020automatic}. Among the many applications of ocean technology, fish monitoring and tracking is a crucial task. Analyzing fish length, width, and mass is essential for maintaining healthy stock in the aquaculture industry, while counting wild fish populations supports marine conservation and coastal reef restoration efforts. Traditional approaches to obtaining such information, such as manual fish measurement, are labor-intensive and inefficient. Deep learning-based segmentation models, which automatically partition each fish into meaningful pixel regions, offer a promising alternative through dense segmentation masks. However, obtaining accurate segmentation masks of fish from underwater videos is difficult due to the unique challenges posed by underwater scenes: caustics, turbidity, attenuation, and suspended particles cause traditional computer vision models to struggle with accurate fish detection and segmentation \cite{kapoor2025graph}.

\begin{figure}[t]  %
    \centering
    \includegraphics[width=\columnwidth]{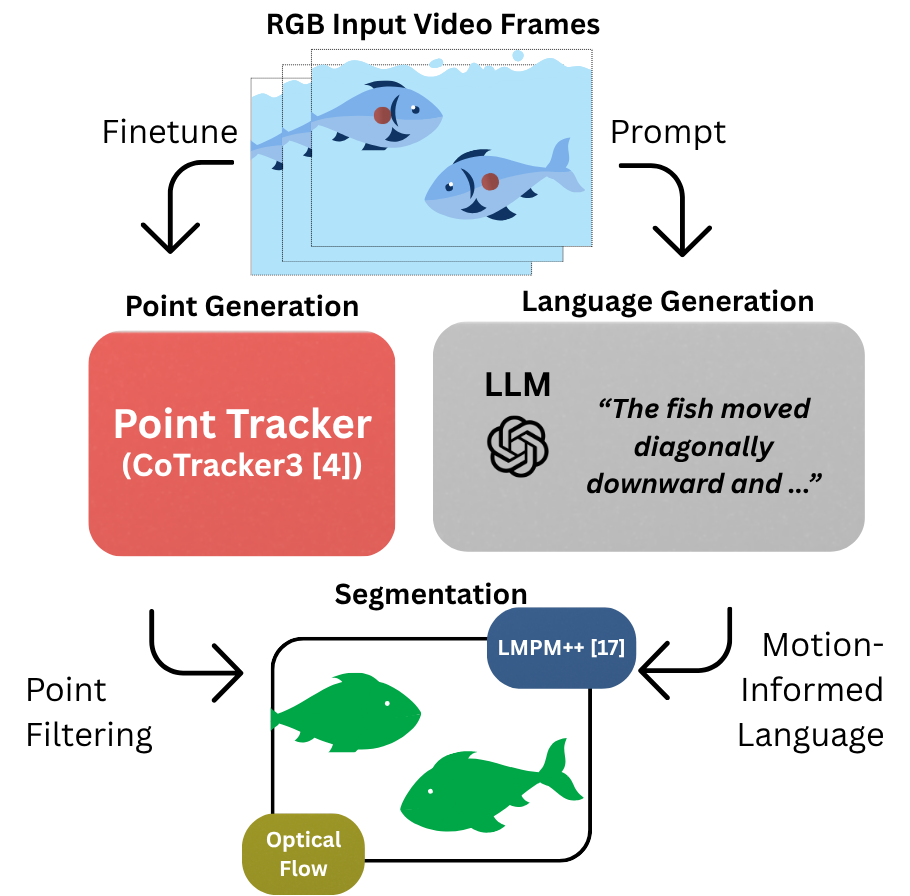}
    \caption{Our proposed method utilizes sparse, easy to annotate point labels for fine-tuning a point tracker and generating novel motion-informed language queries. These guide the creation of dense segmentation masks.}
    \label{fig:pitch_figure}
\end{figure}

Several recent strategies attempt to overcome these challenges. Fully supervised models show robust performance by training on dense segmentation masks, but obtaining per-pixel labels is costly and time-consuming, taking up to two minutes per fish to annotate accurately \cite{saleh2020realistic}. Weakly supervised segmentation models instead require only weak annotations, such as point labels (a single click per fish), drastically reducing labeling effort. However, these models face challenges due to the limited information point labels provide compared to dense labels. We aim to address these limitations and achieve performance comparable to full supervision.

Here, we propose a weakly supervised segmentation pipeline that fully utilizes the information provided by point labels. First, we leverage the strong generalizability of pretrained point trackers, which predict point trajectories across the frames of a video sequence. Using temporal information in a video, point trackers can serve as a general-purpose tracking backbone for larger pipelines. One such tracker, CoTracker3 \cite{karaev2025cotracker3}, has been pretrained on large amounts of synthetic data with point tracks and occlusion annotations. After fine-tuning the tracker with a unique mechanism, we use its predicted points for downstream tasks in our pipeline.

Second, we derive an additional signal from the predicted point labels: a natural language query. We pass the predicted point trajectories into a large language model (LLM) to generate natural language descriptions of the observed motion, capturing behavioral cues such as movement direction, speed, and the object's ``fish'' class. These queries are used to prompt LMPM++, a language-guided motion perception and matching framework \cite{ding2025mevis}, which generates a set of candidate segmentation masks by matching the language description to motion patterns detected across video frames. The candidate masks are filtered using the predicted point labels and further denoised with background subtraction and optical flow \cite{saleh2024track}.

Figure~\ref{fig:pitch_figure} provides an overview of our proposed pipeline, illustrating how point labels are used both to fine-tune a point tracker and to generate motion-informed language queries that jointly guide segmentation. To the best of our knowledge, we are the first to use LLM-generated motion descriptions derived from point trajectories to guide candidate mask generation for weakly supervised fish segmentation. Our pipeline leverages temporal information and introduces a new inference signal through motion descriptions, yielding a robust, weakly supervised pipeline for underwater scenes. In summary, we present the following contributions:

\begin{enumerate}
\item A novel weakly supervised fish segmentation method that requires only point labels for training and utilizes temporal information in videos, eliminating the need for dense per-pixel annotation.
\item A new language-guided framework in which an LLM converts predicted point trajectories into motion-informed natural language queries, introducing behavioral and motion cues as an additional supervisory signal at inference.
\item Experiments on the publicly available DeepFish dataset \cite{saleh2020realistic} to provide qualitative and quantitative comparison to baseline methods.
\end{enumerate}

\section{Related Works}
\subsection{Fish Segmentation Methods}

Although several fully supervised fish segmentation methods have shown robust performance, they require costly dense segmentation labels for training. In contrast, weakly supervised methods are attractive due to the low cost of their training labels (one click per fish for point labels). Affinity-LCFCN \cite{laradji2021weakly} combines a convolutional neural network with a random-walk affinity matrix and trains on point labels using the LCFCN loss. Foundation models such as SAM \cite{kirillov2023segment}, SAM2 \cite{ravi2024sam}, and their fine-tuned underwater variants, AquaSAM \cite{xu2023aquasam} and WaterSAM \cite{hong2024watersam}, are trained on immense datasets, providing zero-shot capabilities when given weak prompts (points, bounding boxes). Other variants, such as LangSAM \cite{medeiros2025langsam}, connect a GroundingDINO detector to SAM to generate segmentation masks from simple language prompts. Although these approaches are label-efficient, none fully address the domain-specific challenges of underwater fish segmentation. Our pipeline is the first to utilize point labels not only for localization, but also to provide behavioral and positional context through language as an additional signal at inference, producing highly accurate segmentation masks that meet or surpass baselines.

\subsection{Point Tracking}
Point tracking aims to predict the locations of a set of query points (2D coordinates on one frame of a video) across all other frames. This differs from classical object tracking, which typically tracks bounding boxes or segmentation masks. Optical flow is a related technique that computes a dense field of pixel displacements between pairs of frames \cite{teed2020raft}; point tracking, in contrast, predicts long-range trajectories of query points across the entire video. Trackers also predict occlusion, determining whether each point is hidden or visible in a given frame. This allows points to be reasoned over: a point that disappears can be recovered later in the video.

Several point tracking methods have been introduced in recent years. Persistent Independent Particles (PIPs) pioneered long-range point tracking \cite{harley2022particle}, treating each pixel as a particle with a multi-frame trajectory. However, PIPs tracks points independently, ignoring correlations between neighboring points, and process videos sequentially, creating a computational bottleneck. TAPIR later combined global per-frame matching for coarse initialization with PIPs-style local iterative refinement, achieving substantially stronger performance \cite{doersch2023tapir}, but it also tracks points independently. CoTracker3 \cite{karaev2025cotracker3} addresses this limitation by jointly tracking multiple points, modeling their correlations through a transformer architecture with specialized time and cross-track attention layers. By attending over both the temporal evolution of individual tracks and the relationships between tracks, CoTracker3 leverages contextual information from neighboring points to recover trajectories through occlusions and out-of-view regions.

We employ CoTracker3 as the point-tracking backbone of our pipeline due to its strong generalization to challenging visual domains. CoTracker3 is pretrained on TAP-Vid-Kubric \cite{doersch2022tap}, a synthetic dataset of more than 1,000 videos annotated with point tracks and occlusion labels. Joint tracking and pretraining on occlusion-annotated synthetic data equip it to handle, to some extent, the occlusions and visual noise typical of underwater scenes. We fine-tune CoTracker3 on underwater fish videos to further adapt it to our domain, and its predicted point trajectories serve as the primary weak signal driving the remainder of the pipeline: they are fed into an LLM to generate motion descriptions and used to filter candidate segmentation masks produced by Language-guided Motion Perception and Matching (LMPM++) \cite{ding2025mevis}.

\subsection{LMPM++}
Language-guided Motion Perception and Matching (LMPM++) \cite{ding2025mevis} is a referring video object segmentation framework proposed as a benchmark method for the MeViS dataset \cite{ding2023mevis}. LMPM++ is trained on MeViS, which contains over 33,000 human-annotated motion expressions in text and audio, with a vocabulary focused primarily on motion. Unlike prior video segmentation methods, LMPM++ is explicitly designed to handle motion-centric language expressions that cannot be resolved from a single frame alone.

LMPM++ first generates language-conditioned queries to detect potential target objects across video frames, representing each candidate as a compact object embedding. This design enables efficient processing of sequences up to 200 frames by feeding object tokens, rather than dense frame features, into a large language model that reasons over the full temporal context to identify and count target objects. To further improve temporal understanding, LMPM++ introduces a temporal-level contrastive loss that trains the model to distinguish motion sequences of different temporal orders, resolving ambiguities such as ``first jumping high then jumping far'' versus ``first jumping far then jumping high.''

In our pipeline, LMPM++ serves as the candidate mask generator. Rather than a human-annotated language expression, we supply an LLM-generated motion description derived from predicted point trajectories, capturing behavioral cues such as movement direction, speed, and object class. LMPM++ uses this query to produce a set of candidate segmentation masks by matching the motion description against motion patterns detected across frames. These candidates are subsequently filtered using our tracked points and refined through background subtraction and optical flow.

\begin{figure*}[t]  %
    \centering
    \includegraphics[width=\textwidth]{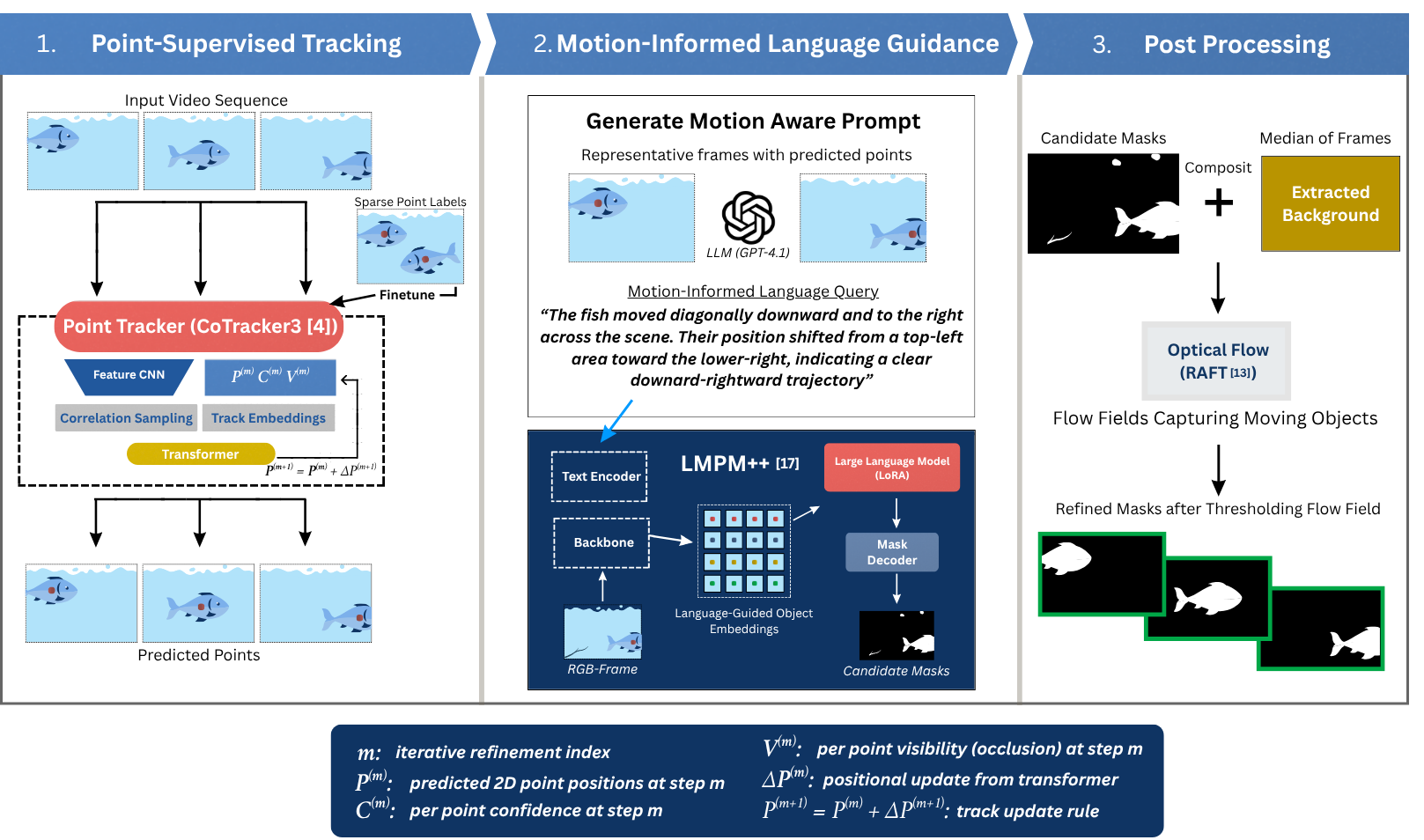}
    \caption{Overview of the proposed pipeline. (1) Point-supervised tracking: a CoTracker3 \cite{karaev2025cotracker3} point tracker, fine-tuned on sparse human-annotated point labels, propagates one point per fish across all frames of the input video, producing predicted point trajectories. (2) Motion-informed language guidance: the predicted trajectories are passed to an LLM, which generates a short natural language description of each fish's motion (e.g., direction and trajectory shape). This description conditions LMPM++ \cite{ding2025mevis}, which encodes the text query and video frames into language-guided object embeddings and decodes a set of candidate segmentation masks. (3) Post-processing: candidate masks are composited onto an estimated background via background subtraction, and RAFT optical flow is computed on the composited frames; only mask pixels exhibiting significant motion are retained, yielding the final refined masks of moving fish.}
    \label{fig:methodology_overview}
\end{figure*}

\section{Methodology}

Figure~\ref{fig:methodology_overview} shows an overview of our proposed method. Our pipeline consists of four main steps:

\begin{enumerate}
\item A fine-tuned point tracker predicts point locations across frames, leveraging the temporal information in video data.
\item An LLM converts predicted point trajectories into natural language queries that capture motion and behavioral cues to aid in fish segmentation.
\item The points and language queries are used to generate and filter candidate masks from LMPM++ \cite{ding2025mevis}, a motion-language guided video object segmentation framework.
\item Optical flow and background subtraction remove noise from the generated segmentation masks.
\end{enumerate}

These components are described in more detail in the following subsections.

\subsection{Fine-Tuning CoTracker3}
We fine-tune CoTracker3 \cite{karaev2025cotracker3} to adapt its pretrained tracking capabilities to the visual characteristics of underwater fish videos while preserving the broad generalization learned from large-scale synthetic pretraining. Specifically, we initialize the model from the pretrained offline checkpoint (CoTrackerThreeOffline) and fine-tune all network parameters except the visibility confidence head.

We select the offline variant because it processes the entire clip jointly, allowing the tracker to reason both forward and backward in time. This is particularly important in underwater scenes, where fish frequently become partially occluded by suspended particles, lighting artifacts, or other fish before reappearing several frames later.

During training, a random temporal offset is sampled from each video. Starting from this offset, we extract a 50-frame clip sampled every 10 frames. Temporally sparse sampling exposes the model to a new temporal slice of each video at every training iteration, improving generalization while keeping the computational load manageable. Because the starting offset is randomized at every iteration, the model eventually observes the full temporal extent of each training video.

The first frame of each sequence is initialized with its human-annotated point labels. For all subsequent frames, the model predicts both the 2D location and the visibility state of every queried point. Fish that leave the frame or become temporarily occluded therefore remain represented in the trajectory, allowing the model to recover them when they reappear.

Training is supervised with two complementary objectives: a trajectory loss for point correspondence and a visibility loss for point visibility. The trajectory loss measures the discrepancy between the predicted and ground-truth positions for all visible points. We adopt the Huber loss, which combines the robustness of L1 with the smoothness of L2 regression: for small errors the loss is quadratic, encouraging precise localization, while for larger errors it transitions to a linear penalty, reducing sensitivity to occasional annotation noise or difficult frames. The trajectory loss is given by:

\begin{equation}
L_{\delta}(y, \hat{y}) =
\begin{cases}
\frac{1}{2}(y - \hat{y})^2 & \text{if } |y - \hat{y}| \leq \delta \\
\delta \left(|y - \hat{y}| - \frac{1}{2}\delta\right) & \text{otherwise}
\end{cases}
\end{equation}

\noindent where $\hat{y}$ and $y$ denote the predicted and ground-truth values, respectively, and $\delta$ is the threshold that controls the transition between the two regimes.

In our setting, where ground-truth annotations are of high quality and sequences are relatively short, $\delta$ is set sufficiently large that most errors fall within the quadratic region. The loss thus behaves similarly to the Euclidean distance (L2) during typical training while retaining robustness to rare mismatches or ambiguities in point tracking.

In addition to each point's location, the tracker predicts whether the point is visible (e.g., not occluded or out of frame). For visibility prediction, we employ the binary cross-entropy loss:

\begin{equation}
L_{vis} = -\frac{1}{N}\sum_{i=1}^{N} \left[ v_i \log(\hat{v}_i) + (1 - v_i) \log(1 - \hat{v}_i) \right]
\end{equation}
\noindent where $\hat{v}_i \in [0,1]$ and $v_i \in \{0,1\}$ denote the predicted and ground-truth visibility of point $i$, respectively, and $N$ is the total number of points.

Because invisible points are less informative than visible ones but still important for long-term tracking consistency, we apply the visibility loss to invisible points with a reduced weight of 0.01. This downweights occluded regions while preserving supervision on occlusion and reemergence events.

The final training objective is a weighted combination of the trajectory and visibility losses:

\begin{equation}
L_{total} = L_{\delta} + \lambda \cdot L_{vis}
\end{equation}

\noindent where $\lambda$ is a scalar weighting factor balancing the two loss terms. We choose $\lambda$ such that trajectory accuracy remains the dominant objective, prioritizing precise point localization while softly encouraging reliable occlusion estimation.

\subsection{Generating Language Queries}
Once the fine-tuned tracker produces point annotations for unseen test sequences, these annotations are used to generate motion-informed language queries. Rather than feeding raw coordinates directly into the segmentation model, we translate each trajectory into a motion-centered text query. This provides a richer supervisory signal: the point trajectory specifies \textit{where} the fish is located, while the language description captures \textit{how} it is moving. For example, two fish may occupy similar regions of the image but move in different directions; a motion description allows the segmentation model to distinguish between them.

For every fish, we generate a language query describing its movement over a 10-frame segment. Descriptions are generated with an LLM (GPT-4.1 \cite{openai2025gpt41}), chosen for its ability to automatically produce controlled, descriptive text. The LLM is provided with the fish's spatial coordinates at three representative frames (first, middle, and last) of each segment, along with the video resolution. Three representative frames are sufficient, as fish motion rarely changes drastically between consecutive frames.

The model is instructed to produce a single 5--10 word description of the fish's positional movement, explicitly excluding numerical values, coordinates, and descriptive adjectives. These compact descriptions preserve the essential motion pattern while remaining general enough to transfer across frames. The prompt is structured as follows:

\begin{quote}
\texttt{``I have a 1920$\times$1080 pixel underwater video clip of fish swimming. Across the clip: Frame 1 of 10: ($x_1$, $y_1$); Frame 5 of 10: ($x_5$, $y_5$); Frame 10 of 10: ($x_{10}$, $y_{10}$). Provide a single 5-10 word description focusing only on the fish's positional movement, with no numbers, coordinates, or extra adjectives.''}
\end{quote}

\subsection{Candidate Mask Generation}
The motion-informed language query is then combined with the corresponding RGB frames and passed into the pretrained LMPM++ model \cite{ding2025mevis}, with the query generated for each 10-frame segment applied to all frames in that segment. Because LMPM++ is trained on MeViS \cite{ding2023mevis}, whose 33,000+ expressions use a motion-focused vocabulary, the LLM-generated queries are well-suited for inference.

LMPM++ generates language-conditioned queries to detect potential target objects across frames, representing each candidate as a compact object embedding. These embeddings are fed into an LLM that reasons over the full temporal context to produce a set of top-$k$ candidate segmentation masks. Rather than relying solely on LMPM++'s language-visual ranking to select the final output mask, we introduce a point-guided filtering step at inference time. For each frame, we retain only the candidate masks that contain at least one tracked fish point, taking the union of all point-containing candidates as the final segmentation output. If no candidate mask contains a point, we fall back to the model's top-ranked prediction. In this way, we leverage temporal information from points through novel motion-informed language prompts while enforcing grounding, ensuring that the final segmentation masks are centered on the predicted point labels.

\subsection{Post-Processing}
After per-frame segmentation masks are generated, a post-processing step applies background subtraction and optical flow \cite{saleh2024track}. This step serves two purposes: (1) refining fish mask boundaries and (2) removing incorrectly segmented inanimate objects. Dense optical flow is well suited to this task, as it estimates per-pixel motion, helping distinguish moving objects from static background regions while improving boundary precision.

First, a rough background of the video sequence is estimated by taking the median of every fifth frame. Each frame's predicted segmentation mask is then used to extract the corresponding foreground region from the RGB frame, which is composited onto the estimated background. The composited frames are passed into RAFT \cite{teed2020raft}, which computes dense flow fields between consecutive frame pairs. Finally, the refined mask retains only pixels that are labeled foreground in the original mask and exhibit significant motion in the dense flow field.

\section{Experiments}
\subsection{Evaluation}
We evaluated our weakly supervised pipeline on both segmentation accuracy and detection capability. For segmentation, we used mean intersection over union (mIoU) to measure the overlap between predicted and ground-truth segmentation masks across all classes. 

We report the performance of our model on the FishSeg test set of the DeepFish dataset \cite{saleh2020realistic}, which contains 186 images from 35 unique video sequences.

\subsection{CoTracker3 Training}
We fine-tuned the pretrained CoTracker3 offline checkpoint on fish videos using AdamW (learning rate $5\times10^{-5}$, weight decay $10^{-5}$) with a OneCycleLR scheduler over 5,000 steps. Each training batch consisted of a single 50-frame video clip sampled at 10-frame intervals, with 256 trajectory points per sample. Gradients were clipped to an L2 norm of 10.0.

Ground-truth point annotations were loaded from a CSV file mapping frame filenames to $(x, y)$ coordinates. For frames lacking exact annotations, the nearest annotated frame within a 50-frame window was used as an estimate. Points were scaled from the original video resolution to the model's input resolution ($384\times512$). Supervision was applied only at annotated frames; unannotated frames contributed zero loss.

\subsection{Comparison with Baselines}
\subsubsection{Baseline Selection}
We compared our pipeline with several state-of-the-art weakly supervised and unsupervised segmentation methods. These included the point-supervised Affinity-LCFCN \cite{laradji2021weakly} and an unsupervised method that generates pseudo-labels from optical flow to train a fully supervised model \cite{saleh2024track}. The latter method does not have publicly available code, so our experiments were carried out on a reimplementation of the original method. Additionally, we compared our approach to methods that leverage foundation segmentation models, including SAM2 \cite{ravi2024sam}, LangSAM \cite{medeiros2025langsam}, and AquaSAM \cite{xu2023aquasam}. Both SAM2 and AquaSAM require ground-truth point prompts at test time, whereas our pipeline requires human-annotated points only during training. LangSAM combines SAM with GroundingDINO to produce segmentation masks from a language prompt.

\subsubsection{Quantitative Results Analysis}
Table~\ref{tab:model_performance} shows quantitative results. Our pipeline surpasses the state-of-the-art baselines in foreground and overall mIoU. Although SAM2 achieves a higher background mIoU, it requires ground-truth point annotations at inference, whereas our method uses human-annotated points only during training to predict point labels for inference. The comparable mIoU between the two methods demonstrates the effectiveness of our approach: our pipeline successfully distills the supervisory signal from sparse training annotations into a fully automatic inference procedure.

\begin{table}[H]
    \centering
    \begin{tabular}{lccc}
        \toprule
        \textbf{Method} & \textbf{Background} & \textbf{Foreground} & \textbf{mIoU} \\
        \midrule
        Affinity-LCFCN \cite{laradji2021weakly} & 0.989 & 0.563 & 0.776 \\
        SAM2\textsuperscript{*} \cite{ravi2024sam} & \textbf{0.996} & 0.758 & 0.873 \\
        LangSAM \cite{medeiros2025langsam} & 0.993 & 0.780 & 0.887 \\
        Saleh et al. \cite{saleh2024track} & 0.992 & 0.818 & 0.905 \\
        Ours & 0.993 & \textbf{0.843} & \textbf{0.918} \\
        \bottomrule
    \end{tabular}
    \caption{Performance comparison of segmentation models. *Requires ground-truth point prompting at test time. Higher is better. \textbf{Bold} is best.}
    \label{tab:model_performance}
\end{table}

\subsubsection{Qualitative Results Analysis}
Figure 3 shows qualitative comparisons on the DeepFish \cite{saleh2020realistic} test set. Affinity-LCFCN captures the general location of each fish but lacks detail around fish borders, producing coarse masks that miss fins and body contours. This likely stems from the fact that its affinity branch is supervised only by the LCFCN loss on point labels, which encourages a single blob per fish and provides no explicit boundary signal. The unsupervised method of Saleh et al. \cite{saleh2024track} produces sharper boundaries but often undersegments, likely because regions with weak apparent motion generate little flow signal and are therefore excluded from the pseudo-labels. LangSAM can incorrectly segment rocks and other background objects by mistaking them for fish, likely because its detector grounds the text prompt on appearance within a single frame, with no motion cue to rule out static look-alikes. In contrast, our method grounds each mask on tracked point labels and conditions LMPM++ on motion-informed language, while optical flow post-processing removes static regions. Together these yield masks that closely follow fish boundaries while suppressing false positives on inanimate objects.

\begin{figure*}[t]  %
    \centering
    \includegraphics[width=\textwidth,height=5cm, keepaspectratio]{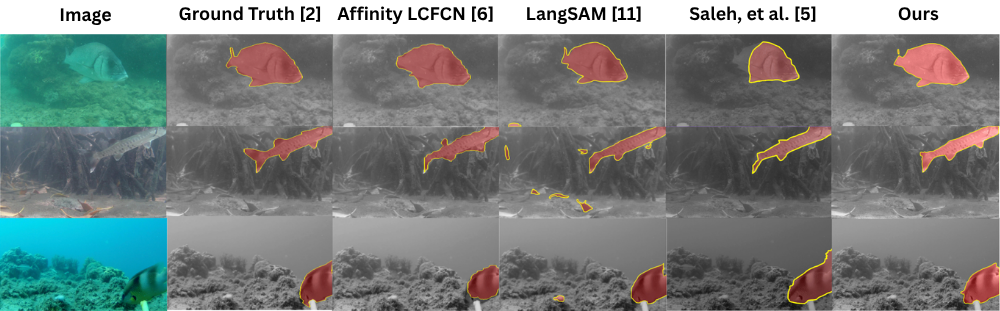}
    \caption{Qualitative comparison on the DeepFish \cite{saleh2020realistic} test set. From left to right: input underwater image, ground-truth segmentation mask, and predictions from Affinity-LCFCN [7], LangSAM [12], Saleh et al. [6], and our pipeline. Each row shows a different scene with varying turbidity, lighting, and fish scale.}
    \label{fig:qualitative}
\end{figure*}

\section{Limitations and Considerations}
Our pipeline involves several computationally intensive components. RAFT optical flow is computed densely across consecutive frame pairs, and LMPM++ \cite{ding2025mevis} processes sequences of up to 200 frames with a large language model backend. Additionally, querying an LLM to generate motion descriptions for every 10-frame segment of every fish adds latency at inference time. Together, these components make the pipeline less suitable for real-time deployment without further optimization. Our pipeline also treats all fish as a single foreground class, making no distinction between species. While this is sufficient for segmentation and counting, applications such as species-level population monitoring would require an additional classification stage not currently present in our framework. Finally, point predictions from the fine-tuned CoTracker3 remain susceptible to the visual noise characteristics of underwater scenes, such as caustics, suspended particles, and turbidity. These can cause points to drift into irrelevant foreground regions, propagating errors into downstream mask generation. Although our point-guided filtering step and background-subtraction post-processing partially mitigate this, sequences with severe visual disturbance remain a challenge. Future work will explore incorporating uncertainty estimates from the tracker's visibility predictions to down-weight unreliable point annotations during mask filtering.

\section{Conclusion}
Underwater fish segmentation remains a challenging problem, complicated by domain-specific visual noise and the high cost of dense pixel-wise annotation. Our work addresses this by constructing a pipeline that extracts maximal supervisory signal from sparse point labels. We fine-tune CoTracker3 \cite{karaev2025cotracker3} to propagate points temporally, convert the resulting trajectories into motion-informed language queries via an LLM, and use these queries to guide candidate mask generation through LMPM++. On the FishSeg benchmark from the publicly available DeepFish dataset \cite{saleh2020realistic}, our proposed approach achieves segmentation performance competitive with state-of-the-art weakly supervised methods without requiring any human input at inference. We hope this work opens a path toward scalable, low-annotation fish analysis tools that can meaningfully support aquaculture management, wild population monitoring, and marine conservation at scale.

\section*{Acknowledgment}

This work was supported by a Stamps Scholarship from the Stamps Scholars Program at the University of Michigan.

\nocite{*}

\bibliographystyle{IEEEtran}
\bibliography{references}

\end{document}